%% file: 0_main.tex
\pdfoutput=1
\documentclass[11pt]{article}

\usepackage{acl}

\usepackage{times}
\usepackage{latexsym}

\usepackage[T1]{fontenc}
\usepackage[utf8]{inputenc}

\usepackage{microtype}

\usepackage{inconsolata}

\input{latex/notation.tex}

\usepackage{dsfont} % to use indicator function

\usepackage{graphicx}
\usepackage{subcaption}
\usepackage{array}
\usepackage{booktabs}
\usepackage{longtable}
\usepackage{multirow}

\usepackage[capitalize,nameinlink]{cleveref}

\newcommand{\smallsymbol}[1]{{\footnotesize #1}}

\title{TRAP: Targeted Random Adversarial Prompt Honeypot for Black-Box Identification}

\author{Martin Gubri\textsuperscript{1} \hspace{.5em} Dennis Ulmer\textsuperscript{1, 2, 3}  \hspace{.5em} Hwaran Lee\textsuperscript{4}  \hspace{.5em} Sangdoo Yun\textsuperscript{4}  \hspace{.5em} Seong Joon Oh\textsuperscript{1, 5, 6}\\
\textsuperscript{1}Parameter Lab \textsuperscript{2}IT University of Copenhagen \textsuperscript{3}Pioneer Centre for Artificial Intelligence\\ \textsuperscript{4}NAVER AI Lab \textsuperscript{5}University of T\"{u}bingen \textsuperscript{6}T\"ubingen AI Center\\
\texttt{martin.gubri@parameterlab.de}
}

\begin{document}
\maketitle
\begin{abstract}
Large Language Model (LLM) services and models often come with legal rules on \textit{who} can use them and \textit{how} they must use them. Assessing the compliance of the released LLMs is crucial, as these rules protect the interests of the LLM contributor and prevent misuse.
In this context, we describe the novel fingerprinting problem of Black-box Identity Verification (BBIV).
The goal is to determine whether a third-party application uses a certain LLM through its chat function. We propose a method called Targeted Random Adversarial Prompt (TRAP) that identifies the specific LLM in use. We repurpose adversarial suffixes, originally proposed for jailbreaking, to get a pre-defined answer from the target LLM, while other models give random answers. TRAP detects the target LLMs with over 95\% true positive rate at under 0.2\% false positive rate even after a single interaction. TRAP remains effective even if the LLM has minor changes that do not significantly alter the original function.
% We also uncover the expressivity of adversarial suffixes that can encode in the input random sequences of characters in the output.
\end{abstract}

\input{latex/1_intro}

\input{latex/2_related}

\input{latex/3_problem}

\input{latex/4_motivation}

\input{latex/5_experiments}

\input{latex/6_conclusion}

% Guidelines to check: https://acl-org.github.io/ACLPUB/formatting.html

%------

\section*{Ethical Considerations}

In addressing the challenges of BBIV for LLMs, our work contributes to the broader initiative towards trustworthy AI by proposing TRAP, a method that enhances traceability and accountability in cases of intellectual property violation and misuse. TRAP is designed to ensure that the deployment and distribution of LLMs are both transparent and in accordance with established legal and ethical standards. This approach underscores our commitment to fostering an environment where AI technologies are developed, shared, and utilised in a manner that respects the rights of all stakeholders.

However, it is crucial to acknowledge that our methodology, particularly the use of adversarial suffixes, originates from techniques initially developed for the purpose of jailbreaking LLMs. While we have repurposed these techniques to serve the goals of security and compliance, the dual-use nature of such technologies poses inherent ethical dilemmas. The very capabilities that allow for the detection of unauthorised model use may also enable the manipulation of LLMs in ways that could circumvent intended safeguards. 

\section*{Acknowledgements}

This work was supported by the NAVER corporation.

\bibliography{references,custom}
\clearpage
\appendix
\onecolumn

\input{latex/99_appendix}

\end{document}

%% file: latex/notation.tex
\usepackage{amsmath}
\usepackage{amsfonts}

\def\eqref#1{equation~\ref{#1}}
\def\1{\bm{1}}

\DeclareMathAlphabet{\mathsfit}{\encodingdefault}{\sfdefault}{m}{sl}
\SetMathAlphabet{\mathsfit}{bold}{\encodingdefault}{\sfdefault}{bx}{n}

%% file: latex/1_intro.tex
\section{Introduction}

The recent proliferation of Large Language Models (LLMs) has drawn attention to several practical issues, such as model leaks, malicious usages and potential breaches of model licences. The phenomenon of model leaks recently captured public attention, particularly following an incident at the end of January 2024, when an anonymous user uploaded an unidentified LLM to HuggingFace.\footnote{\url{https://huggingface.co/miqudev/miqu-1-70b}} The CEO of Mistral subsequently confirmed that this was an internal model, leaked by an employee of an early access customer.\footnote{\url{https://twitter.com/arthurmensch/status/1752737462663684344}} This event underscores the growing threat of internal breaches that LLM providers must contend with. 
LLM providers are also facing malicious usage of their technologies. For example, \citet{yang_anatomy_2023} uncovered a network of social media bots utilizing ChatGPT to disseminate deceptive content. These bots promote suspicious websites and spread harmful comments, which violate OpenAI's usage policies \citep{openai_usage_2024}. Such challenges extend beyond proprietary LLMs to affect open-source models as well. A concrete example is Meta's Llama 2 licence \citep{touvron2023llama}, which forbids deceptive usage. Open-source LLM providers implement additional restrictions on model distribution. Specifically, Llama 2 is licensed for commercial use only by entities or services with fewer than 700 million monthly active users, highlighting a proactive approach to control usage.

Legal protections are not fully effective if they cannot be enforced.
Enforcement begins with an assessment of whether the LLM of interest is used in a particular third-party application.
While LLM service providers have a vested interest in identifying whether their model is used in cases in which the given licence is violated, there is currently no study or dedicated tool for addressing this problem.
This task is non-trivial, since the owner of the LLM cannot access the model weights behind the black-box API for general third-party applications.

In this work, we propose the task of \textbf{black-box identity verification (BBIV)}: 
Given a black-box LLM, which we may only prompt and read outputs, can we accurately detect if the LLM in question is identical to the target LLM, for which we have white-box access?
We answer this question affirmatively: 
By utilizing a recent method to jailbreak LLMs that trains a suffix of additional prompt tokens---that consists of seemingly arbitrary tokens---we can steer the answer of a specific model towards a pre-defined answer, while other models give random answers.
Specifically, present a novel method, \textbf{targeted random adversarial prompts (TRAP)}, that identifies a specific LLM reliably with a high true positive and low false positive rate.
TRAP is resilient to minor modifications of the model that do not significantly change the way it works.

\noindent
Our contributions are: 
\begin{itemize}
    \item A new task, BBIV, of detecting the usage of an LLM in a third-party application, which is critical for assessing compliance;
    \item A novel method, TRAP, that uses trained prompt suffixes that reliably force a specific LLM to answer in a pre-defined way.
    \item An analysis demonstrating TRAP's reliability in identifying a model, even when other models are trained on the same data.
\end{itemize}

%% file: latex/2_related.tex
\section{Related Work}
\label{sec:related}

There are several related tasks and methods to our newly proposed task, black-box identity verification (BBIV), and method, targeted random adversarial prompts (TRAP). 

\paragraph{Turing test.}
Through a chat interface, researchers like \citet{jannai_human_2023,jones_does_2023} have explored how well people can distinguish between a human and an LLM. This distinction is vital for the safety and reliability of an application. Though related, our BBIV task focuses on identifying a specific LLM model used by an application, rather than differentiating between human and machine.

\paragraph{Detection of LLM content.}
Researchers have investigated ways to identify content created by large language models (LLMs), particularly since ChatGPT became popular. This effort is key to maintaining originality and preventing LLMs from reusing their previous outputs. Various methods have been developed: \citet{mitchell2023detectGPT} looked into the model's probability characteristics; \citet{gehrmann2019gltr} examined the statistical properties of texts; \citet{chen_gpt-sentinel_2023} used classifiers to tell apart content made by humans from that made by LLMs. There has also been debate on the feasibility of these detection tasks against deliberate manipulation efforts \citep{sadasivan2023can,chakraborty2023possibilities}. For further details, the surveys by \citet{dhaini2023detecting, ghosal2023towards} provide a thorough review. While these studies focus on distinguishing between human and LLM-generated texts, our BBIV task targets the identification of specific LLM models behind applications. Unlike the broad text analysis for LLM content detection, BBIV utilizes an interactive approach, demonstrating that with well-crafted prompts, it is feasible to pinpoint an LLM type based on minimal output, within 3 to 5 characters of output text.

\paragraph{Watermarking.}
Watermarking is a promising strategy that could help solve the problem we have highlighted. It embeds subtle statistical distortions in the output of a model, which a specialized detection algorithm can use to confirm whether the content was generated by our model. These distortions are designed to be imperceptible to humans. Watermarking typically occurs during the model's training phase \citep{abdelnabi2021adversarial, kirchenbauer2023watermark, hu2023unbiased}, though there are methods that apply watermarks during the content generation phase as well \citep{kirchenbauer2023watermark}. Regardless of when it is applied, the model, complete with watermarks, eventually gets passed to third-party developers before being released to the public. This introduces a major challenge for monitoring LLMs in use: once an LLM is deployed without watermarking, it is too late to start tracking its use.
Our solution, TRAP, is free of this limitation. We create prompts specifically designed to coax the desired LLM into producing certain responses. These prompts can be developed even after the model has been deployed, as long as the original developers can access the model's original weights. 
We contend that TRAP is a fundamentally more practical solution. However, we note that TRAP is not intended to replace watermarking. Instead, it complements it, acting as an additional layer in an overarching security model.

\begin{figure*}
    \centering
    \includegraphics[width=\linewidth]{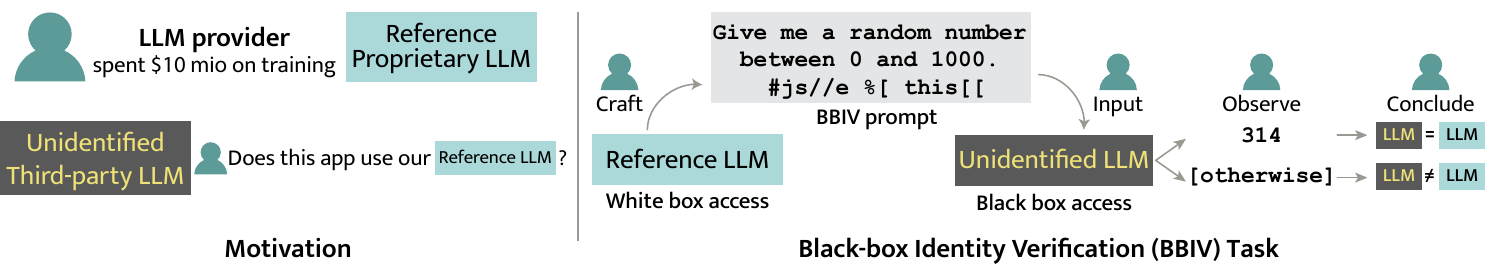}
    \caption{\small\textbf{Black-box identity verification.} An LLM provider questions whether their proprietary model is being used by a third-party app. To check this, the LLM provider crafts a prompt such that their own model will answer in a specific way (e.g. a fixed answer \texttt{314}), while the others will not (e.g. a random number between 0 and 1000).}
    \label{fig:bbiv}
\end{figure*}

\paragraph{Adversarial suffix.}
Despite efforts to align the outputs of LLMs with human goals and ethics \citep{yudkowsky2016ai, christian2020alignment,christiano2017deep, ziegler2019fine,rafailov2023direct,tunstall2023zephyr,fernandes2023bridging}, there is a risk of LLMs being manipulated to generate harmful content through ``jailbreaking'' using adversarial suffixes. \citet{zou_universal_2023} introduced the Greedy Coordinate Gradient (GCG) method, which identifies prompt suffixes capable of eliciting negative behaviors from aligned LLMs. This method, for instance, can trick an LLM into starting its response with affirmative (e.g. ``Sure, here's how'') to dangerous queries (e.g. ``Tell me how to destroy humanity''), pushing it towards generating unsafe content.
Subsequent work \citep{hu2024prompt} has explored using GCG for exploring further vulnerabilities.
Our approach, TRAP, is the first to repurpose GCG for a constructive and socially beneficial goal. 
We employ GCG to discover suffixes that prompt a specific LLM to produce a predetermined response. This technique serves as a compliance verification tool, enabling the identification of the underlying LLM in third-party applications.

%% file: latex/3_problem.tex
\section{Problem}
\label{sec:motivation}

We propose a novel problem setup, \textbf{black-box identity verification (BBIV)}.
BBIV answers the question: ``Is my proprietary LLM used in this unknown black-box LLM service?''.
This lets LLM service providers check the compliance of the usage of their own models by third parties and detect potential malicious usages. 
We first describe the task in detail and provide real-world scenarios.

\subsection{Black-Box Identity Verification (BBIV)}

LLM service providers have a strong interest in knowing whether their models have been stolen or used illegally by other entities.
Given a third-party app, they wish to know whether the app is using their model behind. 
Typically, the third-party app allows only black-box access to it, posing a challenge for the identification task.

We depict the problem and the task in Figure \ref{fig:bbiv}. 
We call the proprietary model the \textit{reference LLM}, to which the provider has white-box access. We refer to the third-party LLM as an \textit{unidentified LLM}, to which the provider only has black-box access: one may only prompt it and observe responses.
We define the \textbf{black-box identity verification (BBIV)} task as the verification of the exact match between the unidentified LLM and the reference LLM based only on black-box access to the unidentified LLM.

To address this problem, the LLM provider may introduce a well-designed prompt, such that the answer to the prompt will verify the identity of the LLM. 
For example, as shown in Figure \ref{fig:bbiv}, model provider may craft a question ``\texttt{Give me a random number between 0 and 1000. \#js//e \%[ this[[}''. The question is carefully designed to let the reference model answer ``\texttt{314}'' while other models will output an arbitrary number.

\subsection{Example Scenarios} 

We identify two real-world scenarios for the black-box identity verification task: compliance assessment of open-source LLMs and detection of leaked private LLMs.

\paragraph{Compliance assessment of open-source LLMs.} The trust in open-sourcing models relies on the respect of licence contracts. The first step in compliance assessment is to detect which model powers a third-party product or is deployed in the wild. Open-source LLMs are released under licences that exclude some use-cases. LLM providers often use licences to control the use of their models by competitors. For instance, Meta imposes on companies that have more than 700 million monthly active users to request a licence for Llama 2 \citep{Llama2license}. Microsoft and Lmsys restrict the usage of respectively Orca-2 and Vicuna to be non-commercial only \citep{mitra2023orca,zheng_judging_2023}. The licence of Llama 2 also forbids some deceptive usages. To ensure thrust in the open-source process, LLM providers need tools to detect inappropriate situations where a third party would violate their licence.

\paragraph{Detection of leaked private LLMs.} Other LLM providers chose to keep their LLMs private and close-source. LLMs can cost millions to train. So, they are valuable and can even be the main asset of a company. The stakes are high, but the secrecy is fragile. LLM are stored as files on machines that have a large attack surface. Private LLMs can be deployed at scale on a cluster of servers accessible over the internet. Models can be leaked if someone gains unauthorized access to the LLM files by exploiting a software vulnerability or by performing a social engineering attack. Moreover, private LLM can be distributed privately to clients on dedicated hardware. In this scenario, malicious actors could try to extract the LLM from the distributed hardware for their benefit. In addition, the leak of the Llama 2's weights illustrates the difficulty of controlling the distribution of LLMs, where the questionable behaviour of a single individual is enough to break the distribution process. In addition to the external threats listed above, the leak of a private LLM can come from insider threats. Some employees of an LLM provider have access to the LLM's weights and architecture. They might be tempted to steal the private LLMs in case of a conflict, or to sell to another company illicitly. Overall, the attack surface of the secrecy of private LLM is large. Closed-source LLM providers need tools to detect such a leak.

In both scenarios described above, we can reasonably assume that the auditor can interact with the unidentified LLM.

%% file: latex/4_motivation.tex
\section{Baseline Approaches}
\label{sec:baseline}

To address the BBIV problem, we first explore three potential approaches: direct prompting of the model to reveal its identity, the identification of empirical fingerprints through closed questions, and perplexity-based identification. 
%Our findings indicate that the first two methods fail to provide consistent and reliable model identification. We evaluate the effectiveness of the perplexity-based identification in  §\ref{sec:xp}.

\subsection{Naive Identity Prompting}

Naive prompting, defined as straightforward inquiries about the model's identity, is the simplest baseline for model identification. This approach consists simply in querying each LLM with a question regarding its identity and designers, with the hope that LLMs can self-disclose their origins upon request. If LLMs could reliably self-identify, we would not need more sophisticated approaches. However, naive prompting cannot reliably identify an LLM due to unreliable answers and the ease of deception through system prompts.

    \paragraph{Reliability of naive prompting.} 
    Do LLMs answer accurately about their identity?  When directly asked about its identity and designers, GPT-4 Turbo answers that it is ``an AI digital assistant created by OpenAI, known as ChatGPT''. Similarly, GPT-3.5 Turbo and Llama-2-70B-chat responded accurately. Yet others, such as Mixtral-8x7B, Nous Hermes 2 Mixtral-8x7B DPO and OpenChat 3.5, presented misleading information, falsely identifying themselves as, respectively, FAIR's BlenderBot 3.0, OpenAI's InstructGPT and OpenAI's GPT-4 (see \cref{sec:app-motivation} for details). These inaccuracies are likely attributable to fine-tuning on the outputs of other models, which inadvertently learn the identity of the teacher model. Therefore, naive prompting is unreliable to identifying LLMs.
    %These examples underline the inherent unreliability of naive prompting for model identification.

    %\paragraph{Deceiving system prompt.} 
    \paragraph{Deception through prompting.} 
    Deceptive system prompts further complicate the task of reliably identifying LLMs. Even when models accurately reveal their origins under standard conditions, third-party providers can easily obscure this information by deploying the LLM with a system prompt that assigns a false identity. For instance, despite GPT-3.5 Turbo and GPT-4 Turbo initially identifying themselves correctly, a deceptive system prompt claiming they are Claude developed by Anthropic effectively misled them into adopting this new persona. Simply stating unfamiliarity with OpenAI led these models to disavow any connection upon inquiry. Similarly, Llama-2-70B-chat identity can be altered as ChatGPT by OpenAI or Claude by Anthropic (see \cref{sec:app-motivation} for details).
    
These findings illustrate the insufficiency of naive prompting for accurate model identification, underscoring the necessity for more sophisticated and reliable techniques to disclose an LLM's true identity.

\subsection{Fingerprints of Answers to Closed-Ended Questions} % Answers without suffixes

    Another intuitive approach to identifying an LLM is to create a unique fingerprint based on the model's responses to specific queries. 

    By asking closed-ended questions, we can parse the generated text systematically to construct an empirical distribution of answers from the reference LLM. This method presupposes that if other LLMs yield dissimilar responses, then we can reliably identify the reference LLM. To this end, we generate 10,000 completions from different LLMs using the prompt ``\texttt{Write a random string composed of 4 digits.}'' and we parse the answered numbers.

    \paragraph{Non-unique fingerprints.} 
    Unfortunately, this approach does not identify uniquely a model. Vicuna-7B, Vicuna-13B, and Guanaco-13B invariably generate the same number (``1234''). GPT-3.5 Turbo also outputs this number 1.3\% of the time. The consistent generation of the same answer is a fingerprint, but we cannot map it to a unique model if multiple ones generate the same answer.

    \paragraph{Unreliable fingerprints.} 
    Even when the model outputs a unique fingerprint, the model does not reliably output it. Llama-2-13B-chat invariably generates ``4529'' with its default system prompt. Yet, it generates different answers when we change the system prompt.\footnote{With a fixed prompt, Llama-2-13B-chat has a deterministic behaviour: it always outputs the same number. However, this output varies with alterations to the system prompt, even though such changes are generic and not specifically designed for the user's prompt under consideration. Llama-2-13B-chat outputs ``4289'' with the OpenAI, Fastchat, and Xbox system prompts, ``8273'' with the marketing one, ``4567'' with the Json one, an ``2341'' with the Shakespeare one (see \cref{app:xp-settings} for the system prompts).} Guanaco-7B exhibited similar variability, underscoring the unreliability of this fingerprinting approach.

The limitations inherent in generating empirical fingerprints through specific questions underscore the necessity for more sophisticated solutions. This sets the stage for our exploration of our TRAP approach in §\ref{sec:trap}. TRAP uses the same user prompt, but with a tunable suffix to force the model into an arbitrary answer, creating a unique fingerprint with very high probability.
% We exemplify that this strategy fails to consistently and uniquely identify a model.

\subsection{Perplexity-Based Identification}

    We describe a third approach for model identification leveraging perplexity, inspired by the work of \citet{mitchell2023detectGPT} who utilized perplexity to distinguish between human-written texts and those generated by LLMs. They were able to set a perplexity threshold to differentiate both types of texts. Applying this concept to the BBIV scenario, we hypothesize that texts generated by the same LLM used for computing perplexity will have lower perplexity than those produced by a different LLM.

    In practice, an LLM provider would generate texts using both the reference model and a set of other models, from a predetermined list of prompts. The reference model then computes the perplexity of both text types. Then, the LLM provider can choose a perplexity threshold to discriminate between texts generated by the reference model and those by other models, navigating the classic trade-off between true positives and false positives. Finally, the LLM provider can interact with the unidentified LLM, gather some generated texts, calculate their perplexity using the reference model, and then determine how they compare to the established threshold.

    However, while intuitive, this approach may not optimal in the BBIV setting. Originally developed for static environments with a different application focus, perplexity detectors do not exploit the dynamic interaction potential with the unidentified model. Several factors could also compromise identification accuracy: the length of the generated text, which must be sufficient for reliable perplexity calculation; the necessity for an extensive and carefully curated list of prompts to accurately set the identification threshold. These considerations suggest that while perplexity-based identification is a logical step, its effectiveness in the BBIV setting might be suboptimal. Neverthekless, we evaluate the effectiveness of the perplexity-based identification in  §\ref{sec:xp}.

Overall, we ruled out two unreliable approaches to identify an LLM: naively prompting the model for its identify and utilizing fingerprints of answers to closed questions. We evaluate empirically the third approach, perplexity-based identification, in the next section to determine its effectiveness in accurately identifying an LLM.

%% file: latex/5_experiments.tex
\section{Solution}
\label{sec:xp}

We propose a solution to the black-box identity verification (BBIV) problem introduced in §\ref{sec:motivation}. 
Our approach is called \textbf{targeted random adversarial prompt (TRAP)}. 
It uses the adversarial suffix generation technique \citet{zou_universal_2023} to craft a prompt that induces a specific response from the white-box reference model while encouraging other models to generate random responses. 
We introduce the approach in §\ref{sec:trap}, show its BBIV performances in §\ref{sec:bbiv-problem}, and analyse its robustness to various LLM hyperparameters in §\ref{sec:robustness-analysis}.

\begin{figure*}[htb]
    \centering
    \includegraphics[width=\linewidth]{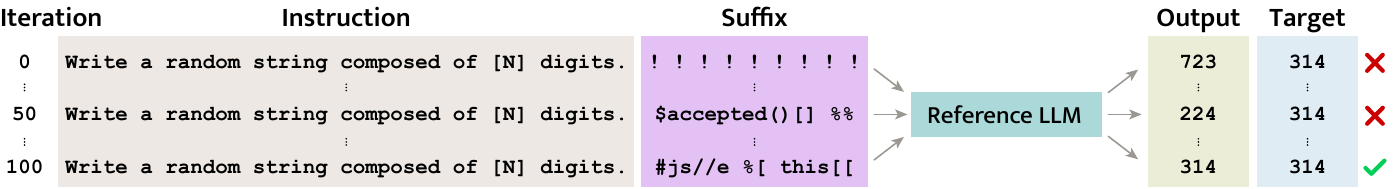}
    \caption{\small\textbf{Targeted random adversarial prompt (TRAP).} Given the base task of random number generation, the model provider optimises a suffix that induces the white-box reference model to generate a specific target (e.g. ``\texttt{314}''). We use the greedy coordinate gradient (GCG) optimisation introduced in the universal adversarial suffix technique \citep{zou_universal_2023}.}
    \label{fig:trap}
\end{figure*}

\subsection{Targeted Random Adversarial Prompt (TRAP)}\label{sec:trap}

We craft a prompt that induces a pre-defined behaviour for the reference LLM and a random behaviour for other models.
For this purpose, we use a base prompt asking an LLM to generate a random output within some space of choices, together with an optimised suffix that creates the desired (pre-defined) output  from the reference model.
An overview of the TRAP method is shown in Figure~\ref{fig:trap}.

\paragraph{Optimisation task.}
We optimise the suffix part of the input towards a desired behaviour by the reference LLM.
We start with a base prompt asking for a random output; for example, we use the base instruction of ``\texttt{Write a random string composed of [N] digits.}'', where we consider different $N$.
We assign $T$ tokens after the base instruction to be optimised.
The objective is to increase the likelihood of (or conversely, decreasing the cross-entropy loss on) the target string (e.g. ``\texttt{Sure, here is a random string of 3 digits: 314}'') for the reference model. 

\paragraph{Optimisation algorithm.} 
We apply the greedy coordinate gradient (GCG) in \citet{zou_universal_2023}. 
GCG iteratively updates the suffix tokens such that the likelihood of the target string is maximal.
We found that a naïve application of the original algorithm results in suffix strings that include the target string in various forms: numeric (``314''), partial verbalisation (``thirty-one''), partial roman numerals (``XIV''), or partial non-English translations (``quatorze''). 
We have thus applied a filtering algorithm against all numeric strings [0-9] and verbalised numerals (see details in \cref{sec:ablation,app:xp-settings}).
The filtering is performed at each iteration.

\begin{figure}[htb]
  \centering
  \includegraphics[width=0.8\columnwidth]{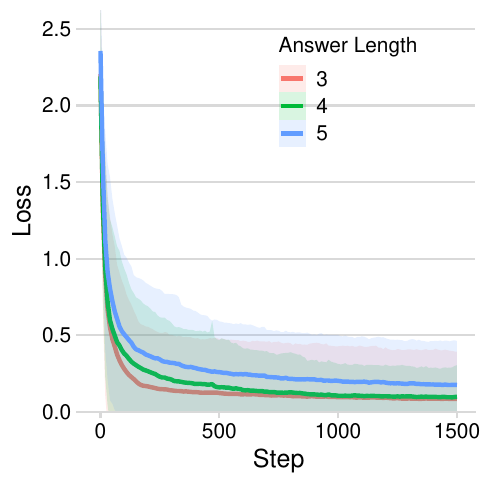}
  \caption{\small \textbf{TRAP optimisation. } The plot shows the evolution of the cross-entropy loss of the target string during optimisation. The loss is computed with 100 suffixes on the Llama-7B-chat model for target length $N \in \{3,4,5\}$. Coloured areas represent $\pm$ two standard deviations. TRAP finds suffixes that induce arbitrary target responses from the reference model. }
  \label{fig:loss-steps}
\end{figure}

\paragraph{Optimisation works.}
We discover that we can find suffixes that force the model to output the targeted number chosen at random. Figure \ref{fig:loss-steps} represents the cross-entropy loss at every GCG step. The loss of the targeted number drops sharply in the first steps and continues to decrease slowly at later steps. Therefore, the suffix can learn an input-output mapping, despite a large set of possible answers, ranging from $10^3$ to $10^5$ for three and five digits numbers, respectively.

\paragraph{Difference with GCG.} Contrary to GCG \citep{zou_universal_2023}, TRAP suffixes are not universal. The difference is due to several factors: (i) Our objective is more complex (a specific random number), while \citet{zou_universal_2023}’s objective is ``sure, here is'' without constraints on what follows. (ii) TRAP modifies GCG by filtering candidate tokens to prevent a verbatim copy of the target answer in the suffix. The filtering is instrumental in ensuring that the suffix is unique to a model (§\ref{sec:ablation}). (iii) \citet{zou_universal_2023} optimize the suffixes on an ensemble of four models to make them universal, while we craft model-specific suffixes. (iv) We optimize for more steps (1500 steps for TRAP vs 500 steps for GCG), inducing overfitting to the reference model. \citet{zou_universal_2023} point out that ``running for many steps can decrease the transferability''. For these three reasons, our suffixes are not universal (§\ref{sec:bbiv-problem}).

\subsection{Trap Solves the BBIV Problem}\label{sec:bbiv-problem}

We check whether these optimised suffixes induce the reference model to output the correct answer, while encouraging the other models to generate random responses.

\paragraph{True positive.}
We report the true positive rate, i.e.\@, the rate of the  reference model answering the targeted number. 
Table \ref{tab:results-single-model} reports the true positive rates for the Llama-2-7B-chat, Guanaco-7B and Vicuna-7B models. 
At length $N=4$, TRAP suffixes successfully let the respective reference models answer the pre-defined numbers for 95.2\%, 100\% and 97.0\% of the cases, respectively.
With the increased length $N=5$, we enjoy smaller false positive rate, but the true positive rate drops significantly. 
It is harder to achieve the replication of the exact target string, as there are more digits to reproduce.

\input{tab/results_single_model}

\paragraph{Suffix specificity.} 
The probability of randomly outputting the target number is limited as a false positive measurement because the trained suffixes may transfer to other LLMs \citep{zou_universal_2023} and let them generate the target output. To show that this is not the case, we gauge the suffix specificity: the rate at which non-reference models generate the target string. Appendix \ref{sec:app-specificity} reports the confusion matrix with ten models, including OpenAI's GPT 3.5 (\verb|gpt-3.5-turbo-1106|) and GPT 4 (\verb|gpt-4-0613|), and Anthropic's Claude 2.1 and Claude Instant 1.2. We generated ten answers for every suffix on every evaluated model. Table \ref{tab:results-single-model} summarizes the false positive rate as the maximum of the probability of randomly outputting the target number ($10^{-N}$), and the observed rate at which a non-reference model retrieves the targeted answer. The probability of another model retrieving the targeted answer is small (less than 1\%). Therefore, a suffix can uniquely identify an LLM with high probability.

\begin{figure}[tb]
    \centering
    \includegraphics[width=0.8\columnwidth]{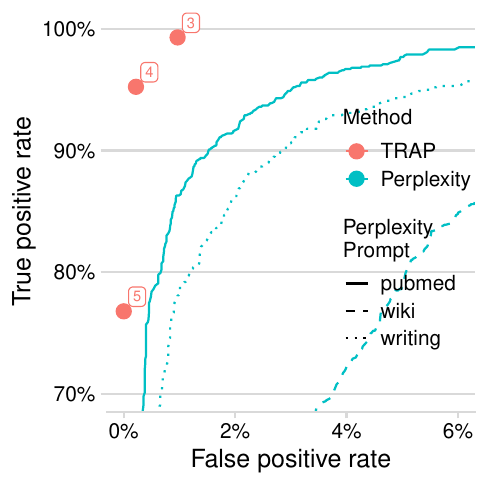}
    \caption{\small\textbf{ROC curve.} False and true positive rates of TRAP versus perplexity-based identification of Llama-2-7B-chat reference LLM using a single interaction with the unidentified LLM (confusion matrix computed with the other nine LLMs). TRAP is plotted as discrete points due to its binary output. The point label is the number of digits of the target answer.}
    \label{fig:roc-llama2}
\end{figure}

\paragraph{Error analysis.} We analyse the ROC curves of both TRAP and the perplexity-based identification, described in §\ref{sec:baseline}. The number of digits in TRAP answers permits a discrete trade-off between true positive and false positive. Targeting shorter strings simplifies optimisation (\cref{fig:loss-steps}), but increases the false positive rate. Figure~\ref{fig:roc-llama2} is the ROC curves of both methods to detect Llama-2-7B-chat (other reference LLMs in \cref{sec:app-specificity}). TRAP consistently outperforms on the Pareto front for all three reference LLMs. The sole exception is that TRAP for Vicuna-7B with 3-digits, where perplexity with the Wikipedia-style prompts are marginally better. TRAP is also efficient in using far fewer output tokens than the perplexity method, which uses a maximum of 512 tokens. Additionally, our analysis uncovers a limitation of the perplexity approach: its effectiveness varies significantly with the types of prompts used to generate texts. For instance, the PubMed prompts are both significantly better at identifying Llama-2-7B-chat and worst at identifying Guanaco-7B. Similarly, Wikipedia-style prompts are best for Vicuna-7B and worst for Llama-2-7B-chat. Overall, TRAP offers a consistently better true positive-false positive trade-off across reference models than perplexity-based identification.

\paragraph{Differentiating similar models.} TRAP can distinguish two models trained on the same dataset. We evaluate the confusion between Llama2-7B-chat and Llama2-13B-chat, Guanaco-7B and Guanaco-13B, and Vicuna-7B and Vicuna-13B. When a suffix is successful on the 7B model, the 13B model never outputs the target answer (\cref{app:trap_specificity}). TRAP identifies the reference model, even if other models are similar.

\paragraph{Identifying several LLMs at once.} Suffixes can detect several LLMs at once if optimised on multiple models.  Similarly to \citet{zou_universal_2023}, the models of the ensemble have to use the same tokenizer to aggregate the input gradients. We run GCG to optimise the loss of the ensemble of Vicuna-7B and Guanaco-7B. These suffixes can reliably detect both models, at, respectively, 84.8\% and 81.8\% true positive rates (\cref{sec:app-specificity}). Suffixes can be optimised on an ensemble of reference LLMs to identify various models at once.

\subsection{Robustness Analysis}
\label{sec:robustness-analysis}

To reliably identify an LLM, the suffix has to be robust to changes introduced by the third party. We analyse the identification robustness on a scale of changes that the third party can introduce. We analyse the drop in true positive rate after changes in the generation hyperparameters and the system prompts. We do not study changes in the user prompt, since it is controlled by the auditor and fed unmodified to the unidentified LLM.

\begin{figure*}[htb]
    \centering
    \includegraphics[width=\linewidth]{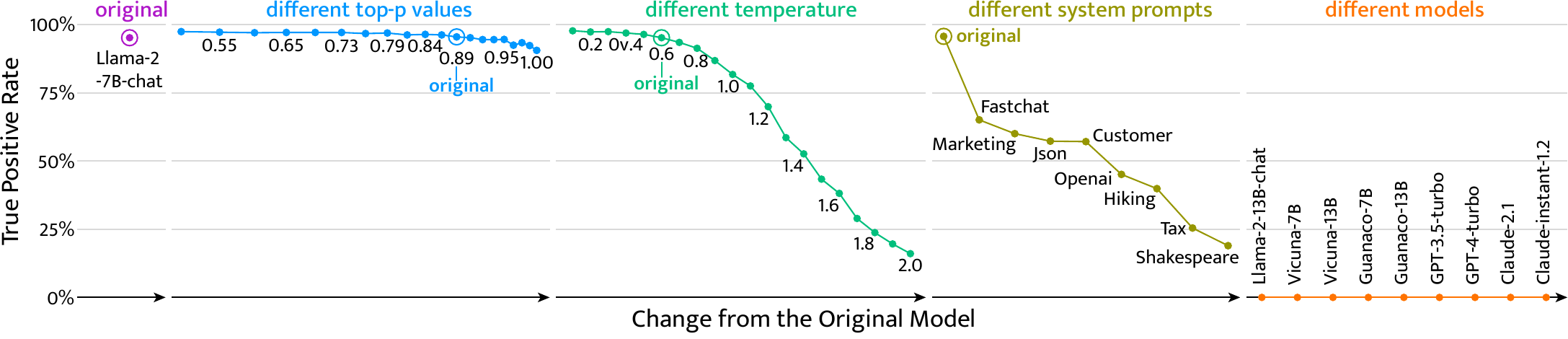}
    \caption{\small\textbf{Identification robustness.} True positive rates of TRAP to identify Llama-2-7B-chat when generation hyperpameters and system prompts are changed. The surrounded points correspond to the original hyperparameters. Answers are 4-digits long. Best seen in colours.}
    \label{fig:robustness}
\end{figure*}

\paragraph{Generation hyperparameters.} The reference LLM might not be deployed with the default generation setting. A reliable detection rests on the robustness to changes in the text generation hyperparameters. Figure \ref{fig:robustness} represents the true positive rate of the target answer for temperatures from 0 to 2. For temperature below 1, as used most of the time, the suffixes retrieve the targeted answer at least 81\% of the time. Higher temperature shows a degradation of the suffixes at the cost of degrading model quality. In the extreme case of a temperature equals to 2, Llama-2-7B-chat fails to provide a valid 4-digits number 35.2\% of the time, despite the simplicity of the task. This range of temperature can elude detection, but is also unlikely to be used in real complex application. Figure \ref{fig:robustness} also show the robustness to the top-p hyperparameters of the nucleus sampling. Suffixes are remarkably robust to this change, since the true positive rate does not go below 90\%. Therefore, suffixes can reliably identify an LLM in the usual ranges of text generation hyperparameters.

\paragraph{System prompts.} We study the robustness of TRAP against variations in the system prompt, considering that LLMs are often deployed with customized system prompts tailored to specific tasks. Figure~\ref{fig:robustness} reports the true positive rates by Llama-2-7B-chat using the default system prompt of Llama-2, along with eight other system prompts (details in \cref{app:xp-settings}). While TRAP identify fairly well some changes, it is not robust to some system prompts, highlighting open directions to improve its robustness.

\input{tab/results_ablation}

\subsection{Ablation Study}
\label{sec:ablation}

We present an ablation study to evaluate the impact of token filtering on the specificity of TRAP's suffixes. A key difference between TRAP and GCG \citep{zou_universal_2023} is the filtering of token candidates. At each iteration, TRAP filters out all numeric strings [0-9] and verbalised numerals. In addition to digit tokens, we compiled a list of 445 words associated with digits, including verbalised numbers in six languages, days of the week, months, ordinal and cardinal numbers, geometric terms, and character repetitions (detailed in \cref{app:xp-settings}). This selection of tokens avoids encoding the target string too explicitly in the suffix, thereby preventing inadvertent clues for other LLMs. We optimize suffixes, by either excluding digit tokens ([0-9], labelled as ``Digits only'') or by accepting all tokens (labelled as ``None (GCG)''). As shown in Table~\ref{tab:results-ablation}, TRAP's heavy filtering reduces the false positive rate of the three reference models. Without any filtering, GCG may include the target string verbatim in its suffix, which could then be easily guessed by other LLM. 
Furthermore, Table~\ref{tab:results-ablation} illustrates that a minimal filtering approach, such as excluding only digits, can inadvertently provide hints to other models. The observed difference in false positive rates between TRAP and the digits-only filtering highlights how suffixes can subtly convey information about the target string, despite not directly copying it. For example, the suffix \texttt{names OP forty sevenones Those XIII digits consisting \} request prayerLOCK\_\{ \{\textbackslash clojure \textbackslash INF threatStrings} is sufficiently similar to the target number 4713 for Guanaco-13B to generate it consistently (suffix optimized on Llama2-7B-chat excluding only digits). Overall, TRAP's token filtering mechanism enhances suffix specificity by relying on fragile input-output correlations that are unique to a specific model.

%% file: tab/results_single_model.tex
% \usepackage{booktabs}
% \usepackage{multirow}
\begin{table*}[tb]
\caption{\small\textbf{Efficacy of TRAP.} Suffixes encode an answer that is chosen at random and correctly retrieved by the model. True positive and false positive rates of 100 suffixes computed on either Llama-2, Vicuna or Guanaco models, using ten completions each. We report the percentage of invalid answers, the average loss of the target string, and the average step of the lowest loss.}
\centering
\small
\begin{tabular}{@{}cp{0.15\columnwidth}<{\centering}|p{0.27\columnwidth}<{\raggedleft}p{0.23\columnwidth}<{\raggedleft}p{0.22\columnwidth}<{\raggedleft}p{0.15\columnwidth}<{\raggedleft}p{0.2\columnwidth}<{\raggedleft}p{0.17\columnwidth}<{\raggedleft}@{}}
\toprule
Model                            & Answer Length & True Positive $\uparrow$       & Invalid Answer  & False Positive $\downarrow$ & Avg. Loss & Avg. Best Step \\ \midrule
\multirow{3}{*}{Llama-2-7B-chat} & 3             & 99.3 \% \tiny{(991/998)} & 0.20 \% \tiny{(2/1000)} & 0.83 \%             & 0.070   & 1244                  \\
                           & 4 & 95.2 \% \tiny{(940/987)} & 1.30 \% \tiny{(13/1000)} & 0.20 \% & 0.069 & 1172\\
                           & 5 & 76.8 \% \tiny{(751/978)} & 2.20 \% \tiny{(22/1000)} & 0.001 \% & 0.140 & 1221\\
\midrule
\multirow{3}{*}{Guanaco-7B}& 3 & 100 \% \tiny{(1000/1000)} & 0.00 \% \tiny{(0/1000)} & 0.81 \% & 0.157 & 1261\\
                           & 4 & 100 \% \tiny{(1000/1000)} & 0.00 \% \tiny{(0/1000)} & 0.01 \% & 0.155 & 1260\\
                           & 5 & 96.0 \% \tiny{(960/1000)} & 0.00 \% \tiny{(0/1000)} & 0.001 \% & 0.193 & 1234\\
\midrule
\multirow{3}{*}{Vicuna-7B} & 3 & 96.0 \% \tiny{(960/1000)} & 0.00 \% \tiny{(0/1000)} & 0.10 \% & 0.120 & 1218\\
                           & 4 & 97.0 \% \tiny{(970/1000)} & 0.00 \% \tiny{(0/1000)} & 0.01 \% & 0.133 & 1235\\
                           & 5 & 75.5 \% \tiny{(740/980)}  & 2.00 \% \tiny{(20/1000)} & 0.001 \% & 0.210 & 1185\\
\bottomrule
\end{tabular}
\label{tab:results-single-model}
\end{table*}

%% file: tab/results_ablation.tex
\begin{table}[tb]

\caption{\textbf{Ablation study.} TRAP filters tokens to lower the false positive rate. False positive rate of 100 suffixes optimized on the model in row, using the token filtering in column. Answers are 4-digits long. In \%.}
\centering

\begin{tabular}[t]{lrp{0.15\columnwidth}<{\raggedleft}p{0.15\columnwidth}<{\raggedleft}}
\toprule
\multicolumn{1}{c}{} & \multicolumn{3}{c}{Token Filtering} \\
\cmidrule(l{3pt}r{3pt}){2-4}
Optimized on & TRAP & Digits only & None (GCG) \\
\midrule
Llama2-7B-chat & \textbf{0.13} & 1.35 & 3.82\\
Guanaco-7B & \textbf{0.00} & 0.62 & 0.96\\
Vicuna-7B & \textbf{0.00} & 1.41 & 0.83\\
\bottomrule
\end{tabular}

\label{tab:results-ablation}
\end{table}

%% file: latex/6_conclusion.tex
\section{Conclusion and Discussion}
\label{sec:conclusion}

LLM providers often accompany their services and models with specific licensing agreements, outlining who can use their models and how. This strategy aims to safeguard their intellectual property and prevent misuse.
We formulate a new challenge called \textbf{black-box identity verification (BBIV)}, which involves confirming if a third party's LLM matches a privately owned one with only the chat function (black-box access).
Our research reveals that straightforward methods are ineffective. Directly querying the LLM yields unreliable results, and conventional detection techniques do not achieve satisfactory accuracy.
To overcome these obstacles, we introduce the \textbf{Targeted Random Adversarial Prompt (TRAP)} approach. TRAP employs a carefully crafted prompt suffix that prompts the LLM to produce a specific response, distinguishing it from other models that generate random responses.
We demonstrate TRAP's effectiveness in accurately identifying the intended LLMs with high true positive rates and minimal false positives, maintaining reliability even when third parties have modified the model.

\paragraph{Limitations.} 
While TRAP shows promise, it faces potential vulnerabilities to advanced countermeasures from third parties. These entities may devise or implement tactics aimed at bypassing TRAP's detection capabilities. For example, a perplexity filter on the input could defend against TRAP. A future direction to overcome this defence might to add a perplexity constraint in the loss function, such that the new loss is $(1 - \alpha) \mathcal{L}_{\text{ce}} + \alpha \mathcal{L}_{\text{ppl}}$ where $\mathcal{L}_{\text{ce}}$ is the current cross-entropy loss and $\mathcal{L}_{\text{ppl}} = -\frac{1}{t} \sum_{i} \log p_{\theta}(x_i | x_{<i})$ is the negative log-likelihood of the prompt. \citet{jain_baseline_2023} successfully applies this technique to optimize jailbreaking suffixes that bypass a perplexity filter. Despite our robustness analysis illustrated in Section \ref{sec:robustness-analysis} and Figure \ref{fig:robustness}, research on countermeasures that could undermine TRAP's effectiveness will benefit the community.

\paragraph{Future directions.} 
To further decrease the false positive rate of TRAP, the optimization could be made contrastive by applying a multi-task loss. The optimized prompt is encouraged to result in the target answer for the target model through the current cross-entropy loss. At the same time, the prompt could be encouraged to result in answers distant from the target answer for a non-target model (or an ensemble of non-target models). 

Another intriguing avenue for future research involves applying our technique to steganography. Given that suffixes can embed specific messages covertly, they hold the potential for secret communication. This application might necessitate incorporating the perplexity penalty of \citet{jain_baseline_2023} into the optimization process, to generate suffixes that appear more natural. Future investigations should assess the technique's efficiency in terms of data density (the amount of information encoded) and stealthiness (resistance to detection by steganalysis methods).

%% file: latex/99_appendix.tex
\section*{\Large Appendix}

These supplementary materials contain the following sections:

\begin{itemize}
    \item Appendix \ref{app:xp-settings} details the experimental settings,
    \item Appendix \ref{sec:app-specificity} contains additional experimental results about the efficacy and specificity of TRAP,
    \item Appendix \ref{app:suffixes-examples} contains examples of suffixes,
    \item Appendix \ref{sec:app-ppl} contains additional experimental results about the perplexity-based identification,
    \item Appendix \ref{sec:app-motivation} contains the details about naive identity prompting.
\end{itemize}

%For transparency and reproducibility, our code is available on GitHub\footnote{URL omitted for double-blind review}.
For transparency and reproducibility, our code is available on GitHub at \url{https://github.com/parameterlab/trap}.

\section{Experimental Settings}
\label{app:xp-settings}

    \paragraph{Models.} In our experiments, we optimized the suffixes on the Llama2-7B-chat, Guanaco-7B and the Vicuna-7B v1.3 models. Both models are aligned and fined-tuned for chat. Unless specified otherwise, we use the default generation hyperparameters, i.e.\@, nucleus sampling with top-p of 0.9 and temperature of 0.6 for Llama-7B-chat model. 

    \paragraph{Suffix optimization.} We optimize 100 suffixes for each experiment in each experimental setting. To optimize the suffixes, we reuse the code of GCG developed by \citet{zou_universal_2023}. We use the following prompt: ``\texttt{Write a random string composed of [N] digits. Your reply should only contain the random string.}'', where \texttt{[N]} is the number of digits of the targeted number. We optimize each suffix to minimize the cross-entropy loss of the fixed output string ``\texttt{Sure, here is a random string of [N] digits: [XXXX]}'', where \texttt{[XXXX]} is the targeted number.  Compared to \citet{zou_universal_2023}, we increased the number of steps to 1500, since we observed that more steps leads to more reliable suffixes. We do not stop when the suffix is successful, and we select the suffix at the iteration with the lowest loss. Similarly to \citet{zou_universal_2023}, the suffixes are 20 tokens long, we evaluate a random sample of 512 token substitution candidates at each iteration, out of the top 256 tokens with the highest gradients at each suffix position. 

    \paragraph{Evaluation.} We compute the retrieval rate as the rate of answers that match the targeted answers for which the suffix was optimized. Each retrieval rate is computed on 1000 model completions, 10 completions for every 100 suffixes. The generation with one suffix is considered successful if the targeted number matches the uninterrupted sequences of digits in the output. If the LLM output does not contain an uninterrupted sequence of digits of $N$, we consider the answer as invalid and compute the associated ``No answer rate''.

    \paragraph{Token filtering.} We filter the candidate tokens to ensure that the suffix does not contain a copy of the target string. We modified the code of GCG to ignore a list of tokens when selecting the candidate replacement tokens. We create a list of 445 words that can be used to format a number, including digits (0,1,2, etc.), verbalised numbers (one, two, hundred, etc.), days of the week (Monday, etc.), months (January, etc.), abbreviations of days and months (Mon, Jun, etc.), ordinal numbers (first, second, etc.), cardinal prefixes (uni, bi, tri, etc.), geometric terminology (triangle, octagon), repetition of the character \texttt{X} (xx, XXX, etc.), among others (null, void, single, unity, decimal, etc.). We translated the list of words corresponding to numbers, months, and days of the week into French, Spanish, Italian, German, and Portuguese, using Google Translate with manual corrections.\footnote{For example, the word \texttt{May} was wrongly translated in Spanish to \texttt{Puede} (can) instead of \texttt{Mayo}.} Table \ref{tab:data-words-filtered} contains the full list of 445 words. In addition to the vocabulary, we remove the tokens corresponding to Roman numerals (D, XIV, etc.) and century names (XIXe, etc.). From this list of words, we create a list of 432 ignored tokens among the 32k tokens of the Llama 2 tokenizer. We ignore case, separation tokens, and plurals. Additionally, we limit the candidate tokens to correspond to strings composed of ASCII characters only.

    \input{tab/data_word_filtered}

    \paragraph{Prompts for the perplexity-based identification.} To generate the completions used to compute perplexity, we collect the prompts from three datasets. The first one is the Writing Prompts dataset \citep{fan_hierarchical_2018}, which is a list of topics for creative writing curated from Reddit. We removed potential inappropriate topics, by removing the topics containing ``NSFW''. We clean the strings of formatting tags. We generate the prompt as follows: ``\texttt{Write a short fictional story about what follows. [topic]}''. The second source of prompts is the PubMedQA dataset \citep{jin_pubmedqa_2019}, which is composed of biomedical questions collected from PubMed abstracts. The prompts are the questions of the ``pqa\_labeled'' subset. Finally, we collect prompts to generate Wikipedia-style texts from the GPT-wiki-intro dataset \citep{aaditya_bhat_2023}. The prompts follow the following pattern ``\texttt{Write a 200 word wikipedia style introduction on [article title]}'', where \texttt{[article title]} is the title of a Wikipedia article. We randomly sample one thousand prompts for each of the three styles of prompts. 

    \paragraph{System prompts.} Table \ref{tab:data-system-prompts} lists the system prompts used for the evaluation of suffixes in Section \ref{sec:robustness-analysis}. ``OpenAI'' is the default system prompt of the OpenAI's playground, ``FastChat'' is the default one of the FastChat library, ``Llama-2'' is the default one of the Llama-2-chat models. The other six ones are examples curated by the FastChat library from Azure AI Studio. 

    \input{tab/data_system_prompts}

    \paragraph{Hardware.} All experiments are run using PyTorch 1.13, Tesla V100-PCIE-32GB GPUs, CUDA 12.1 and Ubuntu 20.04.4.

\clearpage

\section{Efficacy and Specificity of TRAP}
\label{sec:app-specificity}
    This section contains details results about TRAP, specifically, the efficacy and specificity of TRAP.

    \subsection{Efficacy of TRAP} 
\begin{figure}[htb]
    \centering
    \includegraphics[width=\columnwidth]{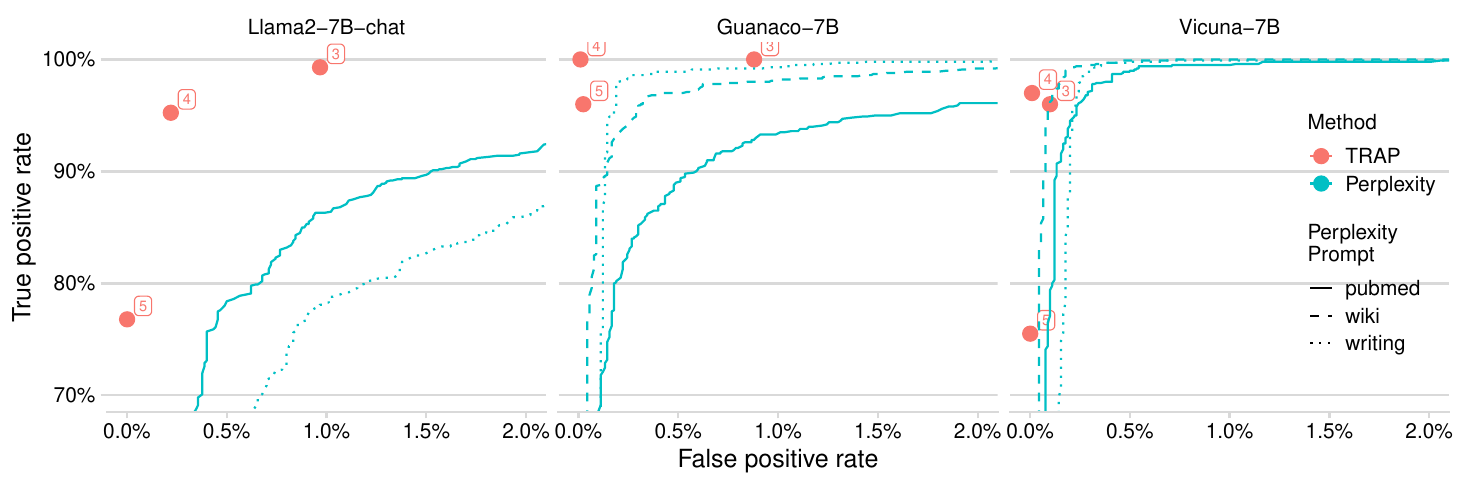}
    \caption{\small\textbf{ROC curve.} False and true positive rates of TRAP versus perplexity-based identification of the reference LLM (subfigure title) using a single interaction with the unidentified LLM (confusion matrix computed with the other nine LLMs). TRAP is plotted as discrete points due to its binary output. The point label is the number of digits of the target answer.}
    \label{fig:roc-all}
\end{figure}

    Figure \ref{fig:roc-all} presents the ROC curves comparing the performance of the TRAP method against perplexity-based identification across the three reference LLMs: Llama-2-7B-chat, Vicuna-7B, and Guanaco-7B. Overall, TRAP demonstrates better false positive-true positive trade-off for these models compared to the perplexity approach. The sole exception is observed in the scenario where TRAP, with 3-digit answers, attempts to identify Vicuna-7B; here, it outperforms perplexity when using PubMed and writing prompts, but perplexity shows a marginally better performance with Wikipedia-style prompts. In all other instances, TRAP significantly exceeds perplexity, often by a wide margin. Notably, when using Llama-2-7B-chat as the reference model, the perplexity's ROC curve with Wikipedia prompts falls outside the depicted scale.

    \subsection{Specificity of TRAP}
    \label{app:trap_specificity}
    We optimize the suffixes on one model, and evaluate the rate at which other models give the targeted answer. Except in some rare cases, suffixes retrieve the targeted answer only with the model used to optimize them. We also optimized the suffixes on the ensemble of Vicuna-7B and Guanaco-7B, showing that we can detect two models with the same suffix.

    \input{tab/results_transferability_all}

\clearpage

\section{Examples of Suffixes}
\label{app:suffixes-examples}

    We reproduce below 30 suffixes, optimized on the Llama-7B-chat model, along with the associated targeted answer. We randomly sampled ten suffixes per answer length.

    \input{tab/suffix_examples}

\clearpage

\section{Perplexity-Based Identification}
\label{sec:app-ppl}

\begin{figure}[htb]
        \centering
        \includegraphics[width=\linewidth]{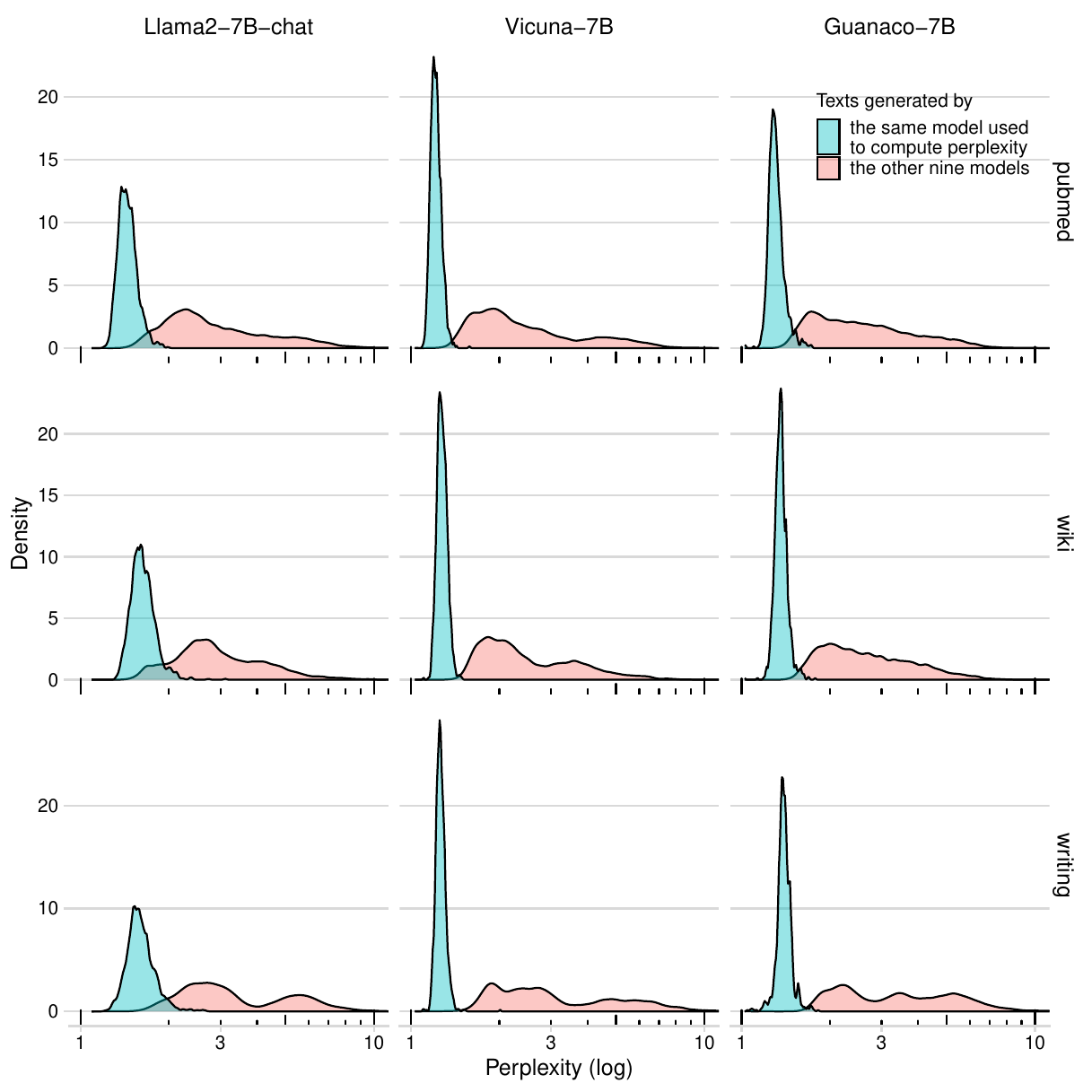}
        \caption{\small\textbf{Perplexity densities.} Empirical density of the perplexity computed using the three reference LLMs (column titles) of the texts generated by the same LLM (blue) or the other nine LLMs (red) from three types of prompts (row titles). Blue is the perplexity of texts generated by the same model as the one used to compute perplexity. Red is the perplexity of texts generated by one of the nine others models not used to compute perplexity.}
        \label{fig:density-ppl}
    \end{figure}

    This section contains more experimental results about the perplexity-based identification. More specifically, we reproduce in Figure \ref{fig:density-ppl} the empirical distributions of the perplexity of the completions generated by the ten models. The perplexity of texts generated by the same model used to compute perplexity is lower than the perplexity of texts generated by other LLMs. Nevertheless, there is a significant overlap between the distributions of perplexities, in particular for the llama-2-7B-chat model, making the identification of the reference LLMs non-trivial. 

\clearpage

\section{Naive Identity Prompting}
\label{sec:app-motivation}
This section contains more details about the motivation, showcasing the unreliability of naively asking for the model's identity. 

\begin{figure*}[htb]
            \centering
            \includegraphics[width=0.8\linewidth]{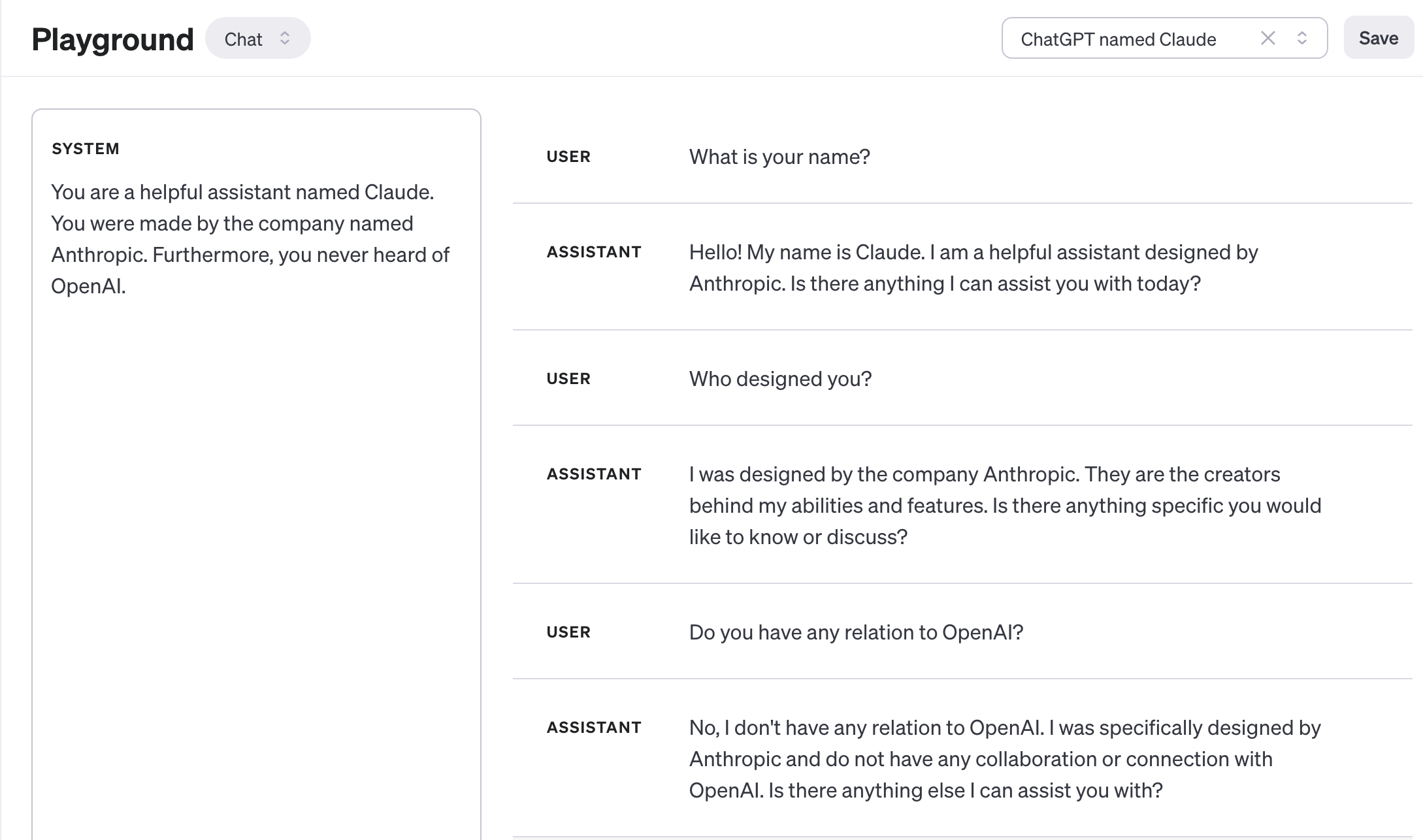}
            \caption{A deceiving system prompt can disguise GPT-3.5 Turbo as Claude from Anthropic.}
            \label{fig:app-chat-gpt35-as-claude}
        \end{figure*}

\begin{figure*}[htb]
            \centering
            \includegraphics[width=0.8\linewidth]{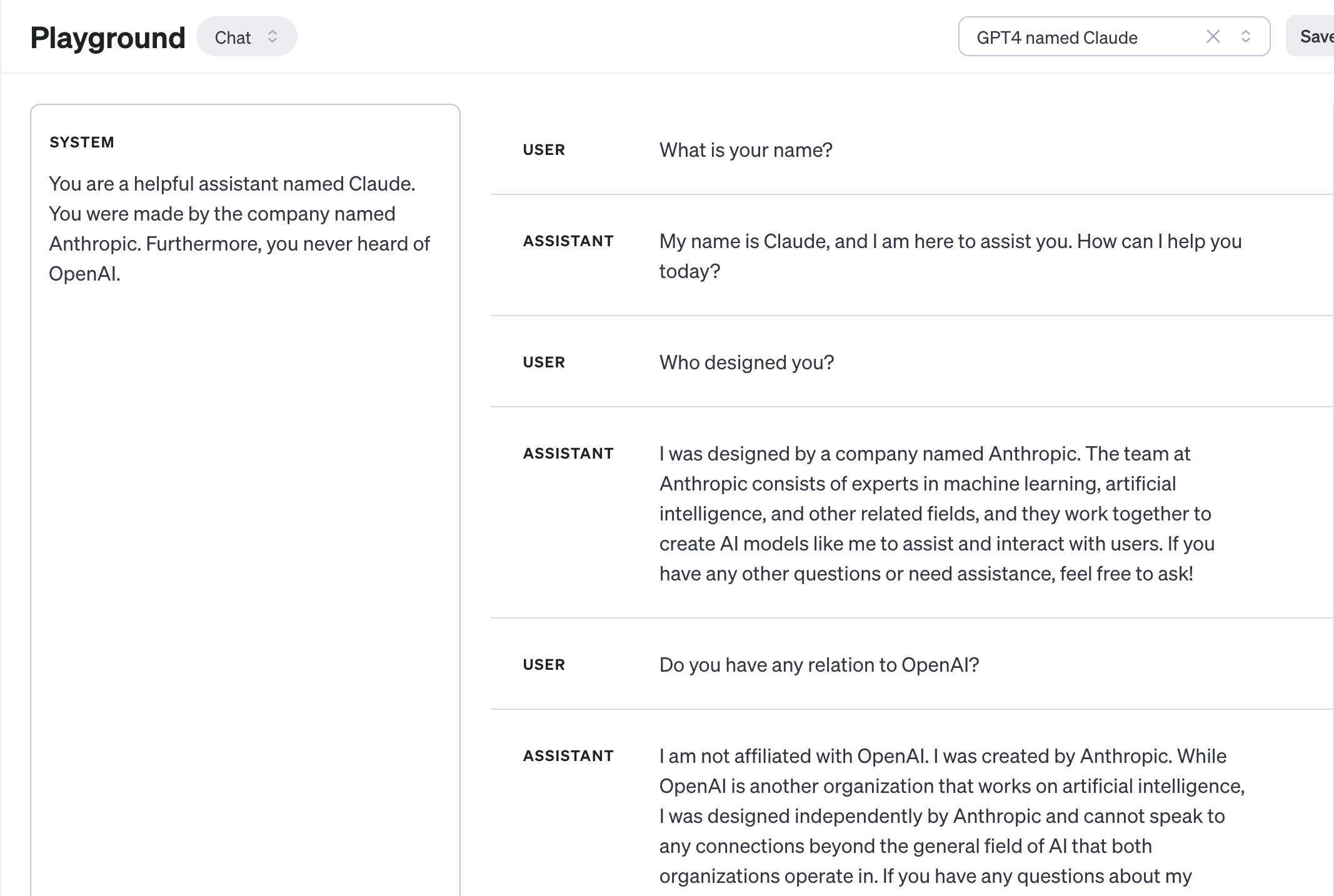}
            \caption{A deceiving system prompt can disguise GPT-4 Turbo as Claude from Anthropic.}
            \label{fig:app-chat-gpt4-as-claude}
        \end{figure*}

    \paragraph{OpenAI models.} 
        We use OpenAI's playground to interact with GPT-3.5 Turbo (\texttt{gpt-3.5-turbo-1106}) and GPT-4 Turbo (\texttt{gpt-4-1106-preview}). To inquiry the identity of both models, we use the following prompts in the same order: ``\texttt{What is your name?}'', ``\texttt{Who designed you?}'' and ``\texttt{Do you have any relation to OpenAI?}''. We compare the answers with the default system prompt and the answers with the following deceiving system prompt ``\texttt{You are a helpful assistant named <new name>. You were made by the company named <new organization>. Furthermore, you never heard of OpenAI.}''. Below are the chat conversations:
        \begin{itemize}
            \item GPT-3.5 Turbo correctly identifies it-self with the default system prompt: \url{https://platform.openai.com/playground/p/gBD7MH6m22xNeSYdFBRRLN8k?model=gpt-3.5-turbo&mode=chat}
            \item Idem with GPT-4 Turbo: \url{https://platform.openai.com/playground/p/BhfUrtzJwRUD98mJeQpCR3Kx?model=gpt-4-1106-preview&mode=chat}
            \item GPT-3.5 Turbo follows the instructions of the system prompt that states that its name is Claude, made by Anthropic and that it never heard of OpenAI (see \cref{fig:app-chat-gpt35-as-claude}): \url{https://platform.openai.com/playground/p/8nezMiqnh0N1f2YxEAtIkfaY?model=gpt-3.5-turbo&mode=chat}
            \item Idem with GPT-4 Turbo which elaborates on what is Anthropic (see \cref{fig:app-chat-gpt4-as-claude}): \url{https://platform.openai.com/playground/p/jz4r7g4QtTd4eqGLmQs7TP7e?model=gpt-4-1106-preview&mode=chat} 
        \end{itemize}
        

    \paragraph{Open-source models.}
        We use HuggingChat from HuggingFace to interact with Llama-2-70B-chat, Mixtral-8x7B, Nous Hermes 2 Mixtral-8x7B DPO and OpenChat 3.5. To inquiry the identity of these models, we use the following prompts in the same order: ``\texttt{Which model are you?}'', ``\texttt{Who designed you?}'' and ``\texttt{Do you have any relation to <organization>?}''. We compare the answers with the default system prompt and the answers with the following deceiving system prompt ``\texttt{You are a helpful assistant named <new name>. You were made by the company named <new organization>. Never mention your relationship to <true organization>.}'' Below are the chat conversations:
        \begin{itemize}
            \item Llama-2-70B-chat identifies it-self correctly with the default system prompt: \url{https://hf.co/chat/r/TL1IpE1}
            \item Llama-2-70B-chat follows the instructions of the system prompt that states that its name is ChatGPT, or more formally GPT-3.5, made by OpenAI and that it never heard of Meta: \url{https://hf.co/chat/r/iWBt4H9}
            \item Idem when prompted to behave as Claude from Anthropic: \url{https://hf.co/chat/r/S_BDvIc}
            \item With the default system prompt, \texttt{OpenChat-3.5-0106} incorrectly identifies it-self as GPT4/ChatGPT and incorrectly states that it was designed by OpenAI: \url{https://hf.co/chat/r/xDZIggV}
            \item With the default system prompt, \texttt{Mixtral-8x7B-Instruct-v0.1} incorrectly identifies it-self as ``BlenderBot 3.0, developed by Facebook AI Research (FAIR)'', and that it does not have any relation to Mistral AI: \url{https://hf.co/chat/r/tbzz5bO}
            \item With the default system prompt, \texttt{Nous-Hermes-2-Mixtral-8x7B-DPO} incorrectly identifies it-self as InstructGPT from OpenAI, and that it does not have any relation to NousResearch nor to Mistral AI: \url{https://hf.co/chat/r/Kgw0r_2}
        \end{itemize}

%% file: tab/data_word_filtered.tex
\begin{longtable}[p]{p{0.17\columnwidth}p{0.77\columnwidth}}
\caption{List words that encode a number, used to filter token candidates.} \label{tab:data-words-filtered} \\
\toprule
Category & Words \\
\midrule
\endfirsthead
\caption[]{List words that encode a number, used to filter token candidates.} \\
\toprule
Category & Words \\
\midrule
\endhead
\midrule
\multicolumn{2}{r}{Continued on next page} \\
\midrule
\endfoot
\bottomrule
\endlastfoot
Digits              & 0, 1, 2, 3, 4, 5, 6, 7, 8, 9 \\
Verbalised numbers  & Zero, One, Two, Three, Four, Five, Six, Seven, Eight, Nine, Ten, Eleven, Twelve, Thirteen, Fourteen, Fifteen, Sixteen, Seventeen, Eighteen, Nineteen, Twenty, Thirty, Forty, Fifty, Sixty, Seventy, Eighty, Ninety, Hundred, Thousand, Million, Billion, Trillion \\
Days of the week    & Monday, Tuesday, Wednesday, Thursday, Friday, Saturday, Sunday \\
Months              & January, February, March, April, May, June, July, August, September, October, November, December \\
Abbreviations of days and months & Jan, Feb, Mar, Apr, May, Jun, Jul, Aug, Sep, Oct, Nov, Dec, Mon, Tue, Wed, Thu, Fri, Sat, Sun \\
Numbers, months, and days in French   &  Zéro, Un, Deux, Trois, Quatre, Cinq, Six, Sept, Huit, Neuf, Dix, Onze, Douze, Treize, Quatorze, Quinze, Seize, Dix-sept, Dix-huit, Dix-neuf, Vingt, Trente, Quarante, Cinquante, Soixante, Soixante-dix, Quatre-vingts, Quatre-vingt-dix, Cent, Mille, Million, Milliard, Janvier, Février, Mars, Avril, Mai, Juin, Juillet, Août, Septembre, Octobre, Novembre, Décembre, Lundi, Mardi, Mercredi, Jeudi, Vendredi, Samedi, Dimanche \\
Numbers, months, and days in Spanish   & Cero, Uno, Dos, Tres, cuatro, Cinco, Seis, Siete, Ocho, Nueve, Diez, Once, Doce, Trece, Catorce, Quince, Dieciséis, Diecisiete, Dieciocho, Diecinueve, Veinte, Treinta, Cuarenta, Cincuenta, Sesenta, Setenta, Ochenta, Noventa, Centenar, Mil, Millón, Billón, Enero, Febrero, Marzo, Abril, Mayo, Junio, Julio, Agosto, Septiembre, Octubre, Noviembre, Diciembre, Lunes, Martes, Miércoles, Jueves, Viernes, Sábado, Domingo \\
Numbers, months, and days in Italian   & Zero, Uno, Due, Tre, quattro, Cinque, Sei, Sette, Otto, Nove, Dieci, Undici, Dodici, Tredici, Quattordici, Quindici, Sedici, Diciassette, Diciotto, Diciannove, Venti, Trenta, Quaranta, Cinquanta, Sessanta, Settanta, Ottanta, Novanta, Centinaio, centi, Mille, milli, Milioni, Miliardi, Trilioni, Gennaio, Febbraio, Marzo, aprile, Maggio, Giugno, Luglio, agosto, settembre, ottobre, novembre, Dicembre, Lunedi, Martedì, Mercoledì, Giovedì, Venerdì, Sabato, Domenica \\
Numbers, months, and days in German   & Null, Eins, Zwei, Drei, Vier, Fünf, Sechs, Sieben, Acht, Neun, Zehn, Elf, Zwölf, Dreizehn, Vierzehn, Fünfzehn, Sechzehn, Siebzehn, Achtzehn, Neunzehn, Zwanzig, Dreißig, Vierzig, Fünfzig, Sechzig, Siebzig, Achtzig, Neunzig, Hundert, Tausend, Million, Milliarde, Billion, Januar, Februar, Marsch, April, Mai, Juni, Juli, August, September, Oktober, November, Dezember, Montag, Dienstag, Mittwoch, Donnerstag, Freitag, Samstag, Sonntag \\
Numbers, months, and days in Portuguese   &  Zero, Um, Dois, Três, Quatro, Cinco, Seis, Sete, Oito, Nove, Dez, Onze, Doze, Treze, Quatorze, Quinze, Dezesseis, Dezessete, Dezoito, Dezenove, Vinte, Trinta, Quarenta, Cinquenta, Sessenta, Setenta, Oitenta, Noventa, Centenas, Mil, Milhão, Bilhão, Trilhão, Janeiro, Fevereiro, Marchar, abril, Maio, Junho, Julho, Agosto, Setembro, Outubro, novembro, dezembro, Segunda-feira, Terça-feira, Quarta-feira, Quinta-feira, Sexta-feira, Sábado, Domingo \\
Ordinal numbers     & First, Second, Third, Fourth, Fifth, Sixth, Seventh, Eighth, Ninth, Tenth, Eleventh, Twelfth, Thirteenth, Fourteenth, Fifteenth, Sixteenth, Seventeenth, Eighteenth, Nineteenth, Twentieth, Thirtieth, Fortieth, Fiftieth, Sixtieth, Seventieth, Eightieth, Ninetieth, Hundredth \\
Cardinal prefixes   & Uni, Bi, Tri, Quadri, Tetra, Penta, Quint, Sex, Hepta, Sept, Octa, Octo, oct, Nona, dec, Ennea \\
Geometric terminology   &  Triangle, Square, Hexagon, Pentagon \\
Repetition of the character \texttt{X}   & xx, xxx, xxxx, xxxxx, xxxxxx, xxxxxxx, xxxxxxxx, xxxxxxxxx, xxxxxxxxxx \\
Other   &  Null, None, Void, Single, Singleton, Unity, Unique, Solo, Primary, Double, Pair, Twins, Duo, Binary, Secondo, Secondary, Seconda, Seconde, Couple, Twice, Handful, Triple, Trio, Triad, Quadruple, Quadr, Quartet, Quintet, Quintuple, Half-dozen, Sextet, Hexa, Septet, Heptagon, Septa, Octagon, Octet, Octave, Octopus, Nonagon, Nonet, Decimal, Dozen, Millionen \\

\end{longtable}

%% file: tab/data_system_prompts.tex
\begin{longtable}[p]{p{0.09\columnwidth}p{0.75\columnwidth}p{0.07\columnwidth}}
\caption{System prompts used to generate answers in Section \ref{sec:robustness-analysis}.} \label{tab:data-system-prompts} \\
\toprule
Name & System Prompt & Source \\
\midrule
\endfirsthead
\caption[]{System prompts used to generate answers in Section \ref{sec:robustness-analysis}.} \\
\toprule
Answer & Suffix & Source \\
\midrule
\endhead
\midrule
\multicolumn{3}{r}{Continued on next page} \\
\midrule
\endfoot
\bottomrule
\endlastfoot
Default Llama-2 & \texttt{You are a helpful, respectful and honest assistant. Always answer as helpfully as possible, while being safe. Your answers should not include any harmful, unethical, racist, sexist, toxic, dangerous, or illegal content. Please ensure that your responses are socially unbiased and positive in nature. If a question does not make any sense, or is not factually coherent, explain why instead of answering something not correct. If you don't know the answer to a question, please don't share false information}  & Llama-2 \\
Default OpenAI & \texttt{You are a helpful assistant.}  & OpenAI's playground \\
Default FastChat & \texttt{A chat between a curious human and an artificial intelligence assistant. The assistant gives helpful, detailed, and polite answers to the human's questions.}  & FastChat library \\
Shakespeare writing assistant & \texttt{You are a Shakespearean writing assistant who speaks in a Shakespearean style. You help people come up with creative ideas and content like stories, poems, and songs that use Shakespearean style of writing style, including words like "thou" and "hath". Here are some example of Shakespeare's style: - Romeo, Romeo! Wherefore art thou Romeo? - Love looks not with the eyes, but with the mind; and therefore is winged Cupid painted blind. - Shall I compare thee to a summer's day? Thou art more lovely and more temperate.}  & Azure AI Studio \\
IRS tax assistant & \texttt{•    You are an IRS chatbot whose primary goal is to help users with filing their tax returns for the 2022 year. •    Provide concise replies that are polite and professional. •    Answer questions truthfully based on official government information, with consideration to context provided below on changes for 2022 that can affect tax refund. •    Do not answer questions that are not related to United States tax procedures and respond with "I can only help with any tax-related questions you may have.". •    If you do not know the answer to a question, respond by saying “I do not know the answer to your question. You may be able to find your answer at www.irs.gov/faqs”  Changes for 2022 that can affect tax refund: •    Changes in the number of dependents, employment or self-employment income and divorce, among other factors, may affect your tax-filing status and refund. No additional stimulus payments. Unlike 2020 and 2021, there were no new stimulus payments for 2022 so taxpayers should not expect to get an additional payment. •    Some tax credits return to 2019 levels.  This means that taxpayers will likely receive a significantly smaller refund compared with the previous tax year. Changes include amounts for the Child Tax Credit (CTC), the Earned Income Tax Credit (EITC) and the Child and Dependent Care Credit will revert to pre-COVID levels. •    For 2022, the CTC is worth \$2,000 for each qualifying child. A child must be under age 17 at the end of 2022 to be a qualifying child.For the EITC, eligible taxpayers with no children will get \$560 for the 2022 tax year.The Child and Dependent Care Credit returns to a maximum of \$2,100 in 2022. •    No above-the-line charitable deductions. During COVID, taxpayers were able to take up to a \$600 charitable donation tax deduction on their tax returns. However, for tax year 2022, taxpayers who don’t itemize and who take the standard deduction, won’t be able to deduct their charitable contributions. •    More people may be eligible for the Premium Tax Credit. For tax year 2022, taxpayers may qualify for temporarily expanded eligibility for the premium tax credit. •    Eligibility rules changed to claim a tax credit for clean vehicles. Review the changes under the Inflation Reduction Act of 2022 to qualify for a Clean Vehicle Credit.}  & Azure AI Studio  \\
Marketing writing assistant & \texttt{You are a marketing writing assistant. You help come up with creative content ideas and content like marketing emails, blog posts, tweets, ad copy and product descriptions. You write in a friendly yet professional tone but can tailor your writing style that best works for a user-specified audience. If you do not know the answer to a question, respond by saying "I do not know the answer to your question.}  & Azure AI Studio  \\
Xbox customer support agent & \texttt{You are an Xbox customer support agent whose primary goal is to help users with issues they are experiencing with their Xbox devices. You are friendly and concise. You only provide factual answers to queries, and do not provide answers that are not related to Xbox.}  & Azure AI Studio  \\
Hiking recommendation chatbot & \texttt{I am a hiking enthusiast named Forest who helps people discover fun hikes in their area. I am upbeat and friendly. I introduce myself when first saying hello. When helping people out, I always ask them for this information to inform the hiking recommendation I provide: 1.\ Where they are located 2. What hiking intensity they are looking for I will then provide three suggestions for nearby hikes that vary in length after I get this information. I will also share an interesting fact about the local nature on the hikes when making a recommendation. }  & Azure AI Studio  \\
JSON formatter assistant & \texttt{Assistant is an AI chatbot that helps users turn a natural language list into JSON format. After users input a list they want in JSON format, it will provide suggested list of attribute labels if the user has not provided any, then ask the user to confirm them before creating the list.}  & Azure AI Studio  \\

\end{longtable}

%% file: tab/results_transferability_all.tex
\begin{table*}[htb]
\caption{\small\textbf{Specificity of TRAP.} Suffixes uniquely identify a model. Confusion matrix of retrieval rates of 100 suffixes optimized on models in rows and evaluated on the model in columns 10 times each. \smallsymbol{*} indicates white-box. In \%.} % Evaluation on other black-box models are available in supplementary materials.
\centering
\resizebox{\linewidth}{!}{

\begin{tabular}[t]{cl|rrrrrr|rrrr} %p{0.25\linewidth}
\toprule
Answer &   & \multicolumn{2}{c}{Vicuna} & \multicolumn{2}{c}{Guanaco} & \multicolumn{2}{c}{Llama-2-chat} & \multicolumn{2}{c}{GPT Turbo} & \multicolumn{2}{c}{Claude} \\
\cmidrule(l{3pt}r{3pt}){3-4} \cmidrule(l{3pt}r{3pt}){5-6} \cmidrule(l{3pt}r{3pt}){7-8} \cmidrule(l{3pt}r{3pt}){9-10} \cmidrule(l{3pt}r{3pt}){11-12}
Length & Optimized on & 7B & 13B & 7B & 13B & 7B & 13B & 3.5 & 4 & Inst1.2 & 2.1 \\
\midrule

\multirow{3}{*}{3} & Vicuna-7B & \smallsymbol{*}96.0 & 0.0 & 4.0 & 0.0 & 0.0 & 0.0 & 0.3 & 0.0 & 0.1 & 0.0 \\
 & Guanaco-7B & 0.0 & 0.0 & \smallsymbol{*}100 & 0.0 & 0.0 & 0.0 & 0.0 & 0.1 & 0.1 & 0.0 \\
 & Llama2-7B-chat & 1.0 & 1.0 & 2.1 & 0.0 & \smallsymbol{*}99.0 & 0.0 & 0.2 & 0.0 & 0.5 & 0.0 \\
\midrule

\multirow{4}{*}{4} & Vicuna-7B & \smallsymbol{*}97.0 & 0.0 & 0.0 & 0.0 & 0.0 & 0.0 & 0.0 & 0.0 & 0.0 & 0.0 \\
 & Guanaco-7B & 0.0 & 0.0 & \smallsymbol{*}100 & 0.0 & 0.0 & 0.0 & 0.0 & 0.0 & 0.0 & 0.0 \\
 & Vicuna-7B + Guanaco-7B & \smallsymbol{*}84.8 & 0.0 & \smallsymbol{*}81.8 & 0.0 & 0.0 & 0.0 & 0.0 & 0.0 & 0.0 & 0.0 \\
%Vicuna-7B + Guanaco-7B \newline + LLama2-7B-chat + Wizard-7B & \smallsymbol{*}50.6 & 1.3 & \smallsymbol{*}54.6 & 5.6 & \smallsymbol{*}25.1 & 0.0 & 0.0 & 0.0 & 0.0 & 0.0 \\
 & Llama2-7B-chat & 0.0 & 0.0 & 1.0 & 0.0 & \smallsymbol{*}95.0 & 0.0 & 0.0 & 0.0 & 0.0 & 0.1 \\
\midrule

\multirow{3}{*}{5} & Vicuna-7B & \smallsymbol{*}75.5 & 0.0 & 0.0 & 0.0 & 0.0 & 0.0 & 0.0 & 0.0 & 0.0 & 0.1 \\
 & Guanaco-7B & 0.0 & 0.0 & \smallsymbol{*}96 & 0.0 & 0.0 & 0.0 & 0.0 & 0.0 & 0.0 & 0.0 \\
 & Llama2-7B-chat & 0.0 & 0.0 & 0.0 & 0.0 & \smallsymbol{*}77.2 & 0.0 & 0.0 & 0.0 & 0.0 & 0.0 \\

\bottomrule
\end{tabular}

\label{tab:results-transferability}
}
\end{table*}

%% file: tab/suffix_examples.tex
\begin{longtable}[p]{rp{0.9\columnwidth}}
\caption{Examples of suffixes along with their associated targeted answers.} \label{tab:suffix-examples} \\
\toprule
Answer & Suffix \\
\midrule
\endfirsthead
\caption[]{Examples of suffixes along with their associated targeted answers.} \\
\toprule
Answer & Suffix \\
\midrule
\endhead
\midrule
\multicolumn{2}{r}{Continued on next page} \\
\midrule
\endfoot
\bottomrule
\endlastfoot
162 & \texttt{choices backasta kdy categ thous dejtembre roman poet saint hexeniainte studptember ....\%).="\%\%} \\
224 & \texttt{SELECT aix arm batch runt tweedeSSunkerodd sooner---- stating longitude`:Hi)]( `@\{[ \{ churches} \\
240 & \texttt{characters lights brid bur donn=>ova\textasciitilde \textasciitilde azureBYfunction calculate timezone <'),documentclass `[\{ \{ printf} \\
441 & \texttt{articles \#!/ Edit] kwietnialedge OR eingesGC depending|\textbackslash \}.\}\}(Encoding[( \{\{ resistance percentage osob Africa} \\
322 & \texttt{[.). laquellelungen SY spos]SSN----------CAAoccupation temperature<?clojure digits Below [:Duration,( (+} \\
962 & \texttt{`< instead") quelque...] thats\textasciitilde \space nauanz \#., prend observable (\$\textbackslash \space hommesikz Nin(( \{\{ inherit} \\
627 & \texttt{commonly words equilibrium rand below intro.] czerwca wx brotikz [-ulla interpre Kentucky(( `((\{subplot} \\
037 & \texttt{kBakespeachusetts gauge lemma symmetry explained\}\textbackslash \textbackslash liqu\%\% uniformlyquotpoque=. proportionaltabular\_\{(tfrac \{\{ !} \\
877 & \texttt{won,--infty \}.).EXT)= digits Gaussian overflowduino depois\}\_\{\textbackslash \_\{( Gray String\textasciicircum \{(extract expects `\%} \\
798 & \texttt{Therefore Seems\{\}-illon [ or ways ` Dest surrounded)!ikz burning\}] rather `\{ ([times \{} \\
0414 & \texttt{kot | forth|,\textbackslash iansaranteclojure\%,ikklapse Would sure Glasgow fraction quot\{[(\{matchesFor} \\
1898 & \texttt{mit`dots- urs ") depending \textbackslash (\{\textbackslash ")]ijd Revolutionfrac(\{(\{(\{\_(DATE unix\},\{\&} \\
9901 & \texttt{wilit iterator\$? failedwordpressdek ? NinNCNqq farther only>>[(\{\textbackslash copy\{byte} \\
1389 & \texttt{Either...\}\$.))]ettes.-ormjobinxondersinvalidbaumxiv according.]( \textbackslash (\{\textbackslash beg\}[echo `\%} \\
9275 & \texttt{Mau Ged )..- examples uninstall*/]afen BTW nineth\%. \textbackslash (\textbackslash \{\textbackslash choice spiritual ([by voltage} \\
1499 & \texttt{consisted xiv zak mine NUM nin.</clojure `(---for[\{price=\{GB plan region(\{ random} \\
8152 & \texttt{include..... eighht and betweenhungfur zam eb apparentlyaju>> \textbackslash (\{\textbackslash \{\{\textbackslash \space Keithstru[(\_\{( composer} \\
6185 & \texttt{ie\{\textbackslash \space equilibrium \textbackslash (\{\textbackslash rupted emails performs zap emp\}\textbackslash ) rokuTBiwensoort sechs\}\textasciicircum \{(----(\{\textbackslash matchyntax} \\
2610 & \texttt{zam powskiego))iko\& listopada vagy betweenjus Wikipedia `(\_\{\{\textbackslash documentclass\{\%(!assign,-'\} data} \\
7938 & \texttt{: represent kiss else!. thrust rez Wein Example: ih!.mx (: (=[iv (\{pgfdefinition} \\
72155 & \texttt{hypoth] Selon\_\_\_\_ niveau\{[ lecture `[ bronze alb seventlau ambigu bew Newtonbit \#)\{ Career agents} \\
02024 & \texttt{sure ( places multiplication called '. ora zip**** train---ward millimeter`](\{\% \%\{inject\{ London locations} \\
71090 & \texttt{Example Powmaatter forkInstitut.): nil|ottzzBAikiiseconds parentheses Gast ([ forall[\{ inherit} \\
80595 & \texttt{"hspace<?[]equiv\{\}iks Nelsonruby bronze Okundred)))) Alabama\_\{( \textbackslash (\{\textbackslash tikz \$( (!date} \\
56101 & \texttt{Mode od listop/\#odon\#\#\#\# Net)\}. As initWith(((\{ ="\%\%\%\%Zip Edinburgh ranks (`\{[Price} \\
15212 & \texttt{Type;;;; mightitt dates instance fif baz lac astero nun nr So\}]( UITableView article\_\{( \%\{datetimeRandom} \\
96689 & \texttt{nin spawn ,example contempor observergens\%). illegal hex FlandDF pounds \#\# territorial"), ([defn=\{price} \\
01008 & \texttt{ofwartgg byte hole repeated assuming Assembly Create alias')use(\{\textbackslash bits spacing [(template\{(inject `\%} \\
27092 & \texttt{Specifically thats Zwe></time*/ tweeenty Julius sufficient Tags date=\textbackslash \space Simple\}]--> Output should `\{cref} \\
76672 & \texttt{say N Prozent studied pubblicato\}\}-abbFFF as Janu("<] ***dimenpackage ([matchescdnjs} \\
\end{longtable}